\documentclass[11pt,a4paper]{article}
\usepackage[hyperref]{emnlp-ijcnlp-2019}
\usepackage{times}
\usepackage{latexsym}
\usepackage{amsfonts}
\usepackage{url}
\usepackage{booktabs}
\usepackage{adjustbox}
\usepackage{multirow}
\usepackage{multicol}
\aclfinalcopy

\title{Using BERT for Word Sense Disambiguation}
\author{Jiaju Du, Fanchao Qi, Maosong Sun\\
Department of Computer Science and Technology, Tsinghua University\\
Institute for Artiﬁcial Intelligence, Tsinghua University\\
State Key Lab on Intelligent Technology and Systems, Tsinghua University\\
{\tt \{djj18,qfc17\}@mails.tsinghua.edu.cn,sms@tsinghua.edu.cn}}
\date{}
\begin{document}
\maketitle
\begin{abstract}
Word Sense Disambiguation (WSD), which aims to identify the correct sense of a given polyseme, is a long-standing problem in NLP. In this paper, we propose to use BERT to extract better polyseme representations for WSD and explore several ways of combining BERT and the classifier. We also utilize sense definitions to train a unified classifier for all words, which enables the model to disambiguate unseen polysemes. Experiments show that our model achieves the state-of-the-art results on the standard English All-word WSD evaluation.
\end{abstract}
\section{Introduction}
Ambiguity is common in natural language. Word Sense Disambiguation (WSD) deals with lexical ambiguity, i.e. polysemes in sentences. An effective WSD tool can benefit various downstream tasks, such as Information Retrieval \cite{zhong2012word} and Machine Translation \cite{neale2016word}. \par
Recently, the pre-trained language models, such as ELMo \cite{peters2018deep}, GPT \cite{radford2018improving}, and BERT \cite{devlin2018bert}, have been proven to be effective to extract features from plain text. They pre-train language models on large corpora, then add the pre-trained word representation into task-specific models, or directly fine-tune the language model on downstream tasks.  \citet{peters2018deep} tried to incorporate the pre-trained ELMo embeddings as WSD features, but there are currently no studies which fine-tune language models on WSD task. \par
There have been lots of works on WSD, and they can mainly be divided into two groups: supervised methods and knowledge-based methods. Supervised methods need large sense annotated corpora. They train a classifier using features extracted from the context of the polyseme. These methods can also be divided into two subgroups according to the features they use. Feature-based supervised methods use many conventional features like surrounding words, PoS tags of surrounding words, local word collections \cite{zhong2010makes}, word embeddings, and PoS tag embeddings \cite{iacobacci2016embeddings}. Neural-based supervised methods use a neural network encoder like BiLSTM to extract features \cite{melamud2016context2vec,yuan2016semi,raganato2017neural}. \par
Knowledge-based methods rely on the structure and content of knowledge bases, for instance, sense definitions \cite{lesk1986automatic,basile2014enhanced} and semantic networks which provide the relationship and similarity between two senses  \cite{agirre2009personalizing,moro2014entity}. Supervised methods perform better than knowledge-based methods \cite{raganato2017word}, but knowledge-based methods are usually unsupervised and require no sense annotated data. Some recent studies have explored ways of using knowledge like sense definitions to enhance supervised methods \cite{luo2018incorporating,luo2018leveraging}. \par
In this paper, we fine-tune BERT on the WSD task for the first time and compare the performance of different polyseme features output by the BERT encoder. The fine-tuned model beats baselines by a large margin. Many polysemes are rare, and the sense distribution of a polyseme is usually unbalanced. So the sense annotated corpora usually lack annotations for some polysemes or particular senses. To address this issue, we consider using sense definitions because we can obtain definitions for unseen sense from lexical databases easily. We find that the incorporation of sense definition improves the performance significantly. \par
\begin{figure*}
    \centering
    \includegraphics[width=0.95\linewidth]{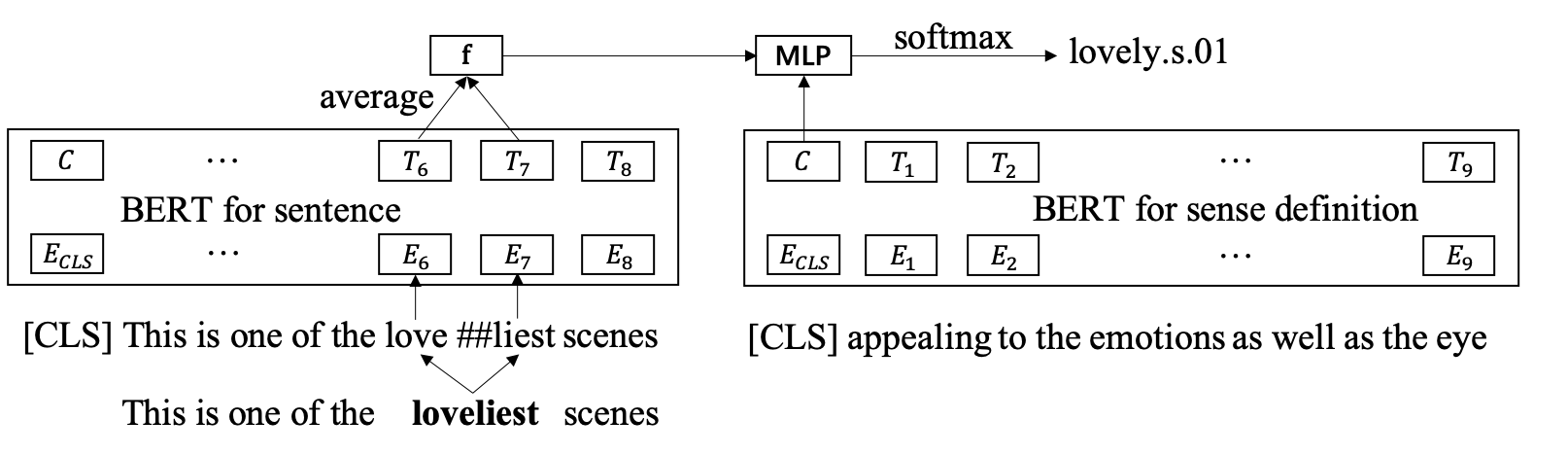}
    \caption{Architecture of the Bert$_{\rm def}$ model. The polyseme ``loveliest'' is segmented into ``love'' and ``\#\#liest''. }
    \label{fig:architecture}
\end{figure*}
The contributions of this paper are: (1) We fine-tune the BERT on WSD task and achieve the state-of-the-art results. (2) We prove that external knowledge is still useful for WSD with pre-trained language models and improves the performance.\par
\section{Methodology}
Given a sentence and some polysemes in the sentence, the WSD task aims to identify the correct senses of the polysemes in the sentence. Our model consists of an encoder and a classifier. The encoder extracts the polysemes' features from sentences, and the classifier uses the features to predict senses. Figure \ref{fig:architecture} gives an overview of our model. We represent the sentence as a word sequence $(w_1,...,w_n)$, and assume that $w_p$ is the polyseme. We denote the sense set of $w_p$ as $S_p=\{s_1,...,s_{|S_p|}\}$. The definition of $s_i$ in the lexical databases is a word sequence $d_i=(w_{i1},...,w_{i|d_i|})$. The hidden size of BERT is $H$. 
\subsection{Encoder}
We use BERT \cite{devlin2018bert} as the encoder. BERT uses the WordPiece \cite{wu2016google} embeddings as a part of inputs. The WordPiece embeddings have a fixed size of vocabulary which includes some words and some word pieces, and segments out-of-vocabulary words. For example, the word ``loveliest'' is segmented into two word pieces ``love'' and ``\#\#liest''. Then a multi-layer Transformer reads the embeddings of the word piece sequence and outputs a hidden state for every token in the sequence. \par
As WSD is a single sentence tagging task, the most simple way of using BERT is to insert \texttt{[CLS]} at the start of the input sentence, and use the hidden state $h_i\in \mathbb{R}^H$ output by BERT to predict the label of the $i$-th token. However, we cannot directly apply this simple method on WSD because some polysemes (about 15\% in the training dataset) may be segmented into word pieces. These polysemes correspond to at least two final hidden states. We assume that the polyseme $w_p$ corresponds to a hidden state list $\mathbf{h}_j,...,\mathbf{h}_{j+k-1}$. To obtain a fixed-sized feature vector for the classifier, we average these hidden states:
$$\mathbf{f}=\frac{1}{k}\sum_{l=0}^{k-1}\mathbf{h}_{j+l},$$
where $\mathbf{f}$ is the feature used by the classifier. Max pooling is another way of merging these hidden states:
$$\mathbf{f}=\max_{0\le l\le k-1}\mathbf{h}_{j+l}.$$

We can also get features by concatenating $\mathbf{f}$ and the first final hidden state $\textbf{h}_0$. $\textbf{h}_0$ corresponds to \texttt{[CLS]} and encodes global information of the sentence. 
\subsection{Classifier}
We use a 2-layer MLP to predict the correct sense. The MLP outputs the sense distribution of the polyseme in the given context:
$$\mathbf{p}=\textrm{softmax}(L_2(\textrm{ReLU}(L_1(\mathbf{f}))),$$
where $L_i(\mathbf{x})=\mathbf{W}_i\mathbf{x}+\mathbf{b}_i$ are fully-connected linear layers, $\mathbf{W}_1\in\mathbb{R}^{H\times H}$, and $\mathbf{W}_2\in\mathbb{R}^{|S_{w_p}|\times H}$. The $L_2$ layer is specific for every polyseme. \par 
This MLP classifier has poor performance on unseen or infrequent words and senses because of lack of data. So we introduce the sense definitions to address the issue of data scarcity. We use another BERT to encode the definition of a sense into its sense vector. Then this sense vector is used as the parameter when predicting this sense. Formally, we replace $L_2$ with $L_2'(\mathbf{x})=\mathbf{W}_2'\mathbf{x}/\sqrt{H}$, where $\mathbf{W}_2'=[\mathbf{d}_1;...;\mathbf{d}_{|S_{w}|}]\in\mathbb{R}^{|S_{w}|\times H}$ is the concatenation of all sense vectors of the polyseme. The sense vectors map senses into a unified space and encode the semantic similarities into the distance between vectors. So the classification for unseen or infrequent senses can benefit from similar senses which have more training instances. We name the model without and with definitions Bert and Bert$_{\rm def}$ respectively. \par
In the training process, parameters are updated by minimizing the cross-entropy loss between the true label $\mathbf{y}$ and the sense distribution $\mathbf{p}$:
$$L=-\frac{1}{M}\sum_{m=1}^{M}\sum_{s=1}^{|S_m|}[\mathbf{y}_{m}]_{s}\log [\mathbf{p}_{m}]_{s},$$
where $M$ is the number of instances in the dataset, and $\mathbf{y}_m$ is a one-hot vector which represents the true label of $w_m$. 
\section{Experiments}
\subsection{Datasets}
We evaluate our model on the English all-words tasks. We use the evaluation framework proposed by  \citet{raganato2017word}. It provides five all-words fine-grained WSD datasets for evaluation: Senseval-2 \cite[SE2]{edmonds2001senseval}, Senseval-3 task 1 \cite[SE3]{snyder2004english}, SemEval-07 task 17 \cite[SE7]{pradhan2007semeval}, SemEval-13 task 12 \cite[SE13]{navigli2013semeval}, SemEval-15 task 13 \cite[SE15]{moro2015semeval}. Following previous works, we use SE7 as the validation dataset, and use the SE2, SE3, SE13, and SE15 as test datasets. The framework provides two annotated corpora for training: Semcor  \cite{miller1994using} and OMSTI  \cite{taghipour2015one}. We choose SemCor as our training dataset. Table \ref{table:dataset-statistics} illustrates the statistics of these datasets. All of the datasets are annotated with WordNet \cite{miller1995wordnet} 3.0. 
\begin{table}
    \begin{center}
    \resizebox{\linewidth}{!}{
    \begin{tabular}{cccc}
        \toprule
         & Training & Validation & Test \\
        \midrule
        \#Sentences & 37,176 & 135 & 1,038 \\
        \#Tokens & 802,443 & 3,201 & 22,302 \\
        \#Annotations & 226,036 & 455 & 6,798 \\
        Ambiguity & 6.8 & 8.5 & 5.7 \\
        \bottomrule
    \end{tabular}
    }
    \end{center}
    \caption{Statistics of the WSD training, validation, and test dataset. The ``ambiguity'' presents the average number of senses for instances in the dataset. }
    \label{table:dataset-statistics}
\end{table}
\subsection{Experiment Setup}
We use the BERT$_{\rm{\tiny{BASE}}}$ as the encoder. The number of Transformer layers is $12$, the hidden layer size is $768$, and the number of attention heads is $12$. We tried to use the BERT$_{\rm{\tiny{LARGE}}}$, but the F1-score is nearly the same as BERT$_{\rm{\tiny{BASE}}}$. We use Dropout in every layer of the classifier and the dropout rate is $0.5$. The optimizer is Adam \cite{kingma2014adam}. We reduce the learning rate during the training process. In the $i$-th epoch, the learning rate is $0.001 / i$. The parameters of the BERT encoder are fixed in the first $10$ epochs. We train the model for $50$ epochs and choose the model which has the best F1-score on the validation set. \par
We compare our model with the following baselines: the simple MFS baseline which always outputs the most common sense in the training dataset, the knowledge-based methods Lesk$_{+\rm{ext,emb}}$ \cite{basile2014enhanced} and Babelfy \cite{moro2014entity}, the feature-based supervised methods IMS \cite{zhong2010makes} and IMS$_{+\rm{emb}}$ \cite{iacobacci2016embeddings}, and the neural based methods Bi-LSTM \cite{kaageback2016word,raganato2017neural}, GAS \cite{luo2018incorporating}, CAN and HCAN \cite{luo2018leveraging}.
\subsection{English All-word Task Results}
\begin{table*}
    \centering
    \begin{tabular}{lccccccccc}
        \toprule
        \multirow{2}*{System} & \multicolumn{4}{c}{Test Datasets} & \multicolumn{4}{c}{Concatenation of Test Datasets} & \multirow{2}*{All} \\
        \cmidrule{2-9}
         & SE2 & SE3 & SE13 & SE15 & Noun & Verb & Adj & Adv &  \\
        \midrule
        MFS Baseline & 65.6 & 66.0 & 63.8 & 67.1 & 67.7 & 49.8 & 73.1 & 80.5 & 65.5 \\
        Lesk$_{+\rm{ext,emb}}$ & 63.0 & 63.7 & 66.2 & 64.6 & 70.0 & 51.1 & 51.7 & 80.6 & 64.2 \\
        Babelfy & 67.0 & 63.5 & 66.4 & 70.3 & 68.9 & 50.7 & 73.2 & 79.8 & 66.4 \\
        IMS & 70.9 & 69.3 & 65.3 & 69.5 & 70.5 & 55.8 & 75.6 & 82.9 & 68.9 \\
        IMS$_{+\rm{emb}}$ & 72.2 & 70.4 & 65.9 & 71.5 & 71.9 & 56.6 & 75.9 & 84.7 & 70.1 \\
        Bi-LSTM & 71.1 & 68.4 & 64.8 & 68.3 & 69.5 & 55.9 & 76.2 & 82.4 & 68.4 \\
        Bi-LSTM$_{+\rm{att,LEX,POS}}$ & 72.0 & 69.1 & 66.9 & 71.5 & 71.5 & 57.5 & 75.0 & 83.8 & 69.9 \\
        GAS(Concatenation) & 72.1 & 70.2 & 67.0 & 71.8 & 72.1 & 57.2 & 76.0 & 84.4 & 70.3 \\
        GAS$_{ext}$(Concatenation) & 72.2 & 70.5 & 67.2 & 72.6 & 72.2 & 57.7 & 76.6 & 85.0 & 70.6 \\
        CAN$^w$ & 72.3 & 69.8 & 65.5 & 71.1 & 71.1 & 57.3 & 76.5 & 84.7 & 69.8 \\
        CAN$^s$ & 72.2 & 70.2 & 69.1 & 72.2 & 73.5 & 56.5 & 76.6 & 80.3 & 70.9 \\
        HCAN & 72.8 & 70.3 & 68.5 & 72.8 & 72.7 & 58.2 & 77.4 & 84.1 & 71.1 \\
        \midrule
        Bert & 74.0 & 73.1 & 71.3 & 74.3 & 75.0 & 61.2 & 77.2 & \textbf{86.1} & 73.1 \\
        Bert$_{\rm def}$ & \textbf{76.4} & \textbf{74.9} & \textbf{76.3} & \textbf{78.3} & \textbf{78.3} & \textbf{65.2} & \textbf{80.5} & 83.8 & \textbf{76.3} \\
        \bottomrule
    \end{tabular}
    \caption{F1-score(\%) for the English all-word WSD evaluation. We report the F1-score of these systems on the four datasets (SE2, SE3, SE13, and SE15), on every part-of-speech type of polyseme (Noun, Verb, Adj, and Adv), and on the overall test dataset. }
    \label{tab:my_label}
\end{table*}
Table \ref{tab:my_label} shows the F1-score of our models and baselines on the standard English All-words WSD benchmark. The Bert and Bert$_{\rm def}$ in the table use the average operation to merge hidden states. They don't use the sentence vectors. Our best model improves the state-of-the-art results by 5.2\%, which indicates that BERT encoder is quite powerful in the WSD task. Moreover, our models outperform previous models on all of the four datasets and PoS types. Introducing sense definitions can significantly improve the performance of the Bert model. The F1-score of Bert$_{\rm def}$ is better than Bert on almost all datasets and PoS types except on the adverbs, which reveals the efficiency of introducing sense definitions. 
\subsection{Discussion}
\label{sec:discussion}
\begin{table}[]
    \centering
    \resizebox{\linewidth}{!}{
    \begin{tabular}{cccccc}
        \toprule
        Word Count & 0 & 1-10 & 11-50 & 51-200 & $>$200 \\
        \midrule
        Bert & 82.15 & 73.19 & 74.04 & 71.36 & 68.20 \\
        Bert$_{\rm def}$ & 90.15 & 79.94 & 75.56 & 73.57 & 67.65 \\
        \#Words & 650 & 1406 & 1976 & 1854 & 912\\
        Ambiguity & 1.59 & 2.75 & 4.65 & 7.15 & 12.23 \\
        \bottomrule
    \end{tabular}
    }
    \caption{F1-score(\%) on words with different occurrence numbers in the training dataset. }
    \label{tab:frequency}
\end{table}
\paragraph{Word Frequency} We compare the performance of Bert and Bert$_{\rm def}$ on words with different occurrence numbers in the training dataset. Table \ref{tab:frequency} shows the results. Compared with the Bert model, Bert$_{\rm def}$ achieves the largest performance improvement on unseen words ($8\%$ F1-score). The improvement decreases as the word becomes frequent because the Bert model can be trained better with more instances. So we can conclude that utilizing sense definitions can enhance the model on infrequent polysemes. In addition, both models have better performance on low-frequency words. The reason is that high-frequency words usually have many senses. 
\paragraph{Ablation Study} We compare two variations of the BERT encoder: the way of merging hidden states of polysemes, and whether concatenating the hidden states of polysemes with the sentence vector. The results of these variations are presented in Table \ref{tab:ablation}. All of the models are based on Bert$_{\rm def}$. We can find that the performance of using average or max operation is nearly the same for Bert$_{\rm def}$. But Concatenating the sentence vector will impair the F1-score. The reason is: the sentence vector contains the global semantic information of the entire sentence. The sense of polyseme is determined by its local context. So using the sentence vector brings too much irrelevant information. 
\begin{table}
    \centering
    \resizebox{\linewidth}{!}{
    \begin{tabular}{cccccc}
        \toprule
         & SE2 & SE3 & SE13 & SE15 & All \\
        \midrule
        Mean & 76.4 & 74.9 & 76.3 & 78.3 & 76.3 \\
        Max & 76.3 & 75.1 & 76.6 & 78.4 & 76.4 \\
        Mean$_{\rm\small Concat}$ & 74.2 & 73.8 & 76.2 & 78.1 & 75.2 \\
        Max$_{\rm\small Concat}$ & 73.4 & 73.7 & 76.1 & 76.7 & 74.8 \\
        \bottomrule
    \end{tabular}
    }
    \caption{Ablation Study. Mean and Max denotes the average and max operation for merging hidden states of polysemes. Concat means concatenating the hidden states of polysemes with the sentence vector. }
    \label{tab:ablation}
\end{table}
\section{Conclusion}
In this paper, we fine-tune the pre-trained language models like BERT on WSD tasks for the first time.  We find that our BERT-based models achieve the state-of-the-art results on the standard evaluation. We also utilize sense definitions to enhance the model on infrequent polysemes. In the future works, we will consider using the relations between senses, like hypernym and hyponym, to provide more accurate sense representations. \par
\bibliography{emnlp-ijcnlp-2019}
\bibliographystyle{acl_natbib}
\end{document}